\documentclass[runningheads]{llncs}
\usepackage{eccv}

\usepackage{eccvabbrv}

\usepackage{graphicx}
\usepackage{booktabs}
\usepackage{wrapfig}
\usepackage[accsupp]{axessibility} 

\usepackage[pagebackref,breaklinks,colorlinks,citecolor=eccvblue]{hyperref}

\usepackage{orcidlink}

\usepackage{multirow}
\usepackage{colortbl}
\usepackage{marvosym}

\begin{document}

\title{AgentVLN: Towards Agentic Vision-and-Language Navigation} 

\titlerunning{AgentVLN: Towards Agentic Vision-and-Language Navigation}

\author{Zihao Xin\inst{1}  \and
Wentong Li\inst{1}*(\Letter)  \and 
Yixuan Jiang\inst{1} \and Ziyuan Huang\inst{1} \and Bin Wang\inst{2} \and \\ Piji Li\inst{1} \and Jianke Zhu\inst{3} \and Jie Qin\inst{1} \and Shengjun Huang\inst{1}(\Letter)}

\authorrunning{Xin et al.}

\institute{Nanjing University of Aeronautics and Astronautics \and Shandong University \and Zhejiang University\\
*Project lead \ \ \ \ \  \Letter Corresponding authors}

\maketitle

\begin{abstract}
  Vision-and-Language Navigation (VLN) requires an embodied agent to ground complex natural-language instructions into long-horizon navigation in unseen environments. While Vision-Language Models (VLMs) offer strong 2D semantic understanding, current VLN systems remain constrained by limited spatial perception, 2D–3D representation mismatch, and monocular scale ambiguity.
  In this paper, we propose AgentVLN, a novel and efficient embodied navigation framework that can be deployed on edge computing platforms. 
  We formulate VLN as a Partially Observable Semi-Markov Decision Process (POSMDP) and introduce a \textbf{VLM-as-Brain} paradigm that decouples high-level semantic reasoning from perception and planning via a plug-and-play skill library.  
  To resolve multi-level representation inconsistency, we design a cross-space representation mapping that projects perception-layer 3D topological waypoints into the image plane, yielding pixel-aligned visual prompts for the VLM.
  Building on this bridge, we integrate a context-aware self-correction and active exploration strategy to recover from occlusions and suppress error accumulation over long trajectories.
  To further address the spatial ambiguity of instructions in unstructured environments, 
  we propose a Query-Driven Perceptual Chain-of-Thought (QD-PCoT) scheme, enabling the agent with the metacognitive ability to actively seek geometric depth information. 
  Finally, we construct AgentVLN-Instruct, a large-scale instruction-tuning dataset with dynamic stage routing conditioned on target visibility.
  Extensive experiments show that AgentVLN consistently outperforms prior state-of-the-art methods (SOTA) on long-horizon VLN benchmarks, 
  offering a practical paradigm for lightweight deployment of next-generation embodied navigation models. Code: \url{https://github.com/Allenxinn/AgentVLN}.
\end{abstract}

\section{Introduction}
\label{sec:intro}
Vision-and-Language Navigation (VLN) requires embodied agents to make sequential decisions in complex physical environments by following natural-language instructions based on  dynamically acquired egocentric observations. 
Despite the strong semantic understanding exhibited by Vision-Language Models (VLMs) on 2D images~\cite{qwen2_vl, qwen3_vl, qwen25_vl, llava,chen2024internvl,shi2024eagle,li2025tokenpacker}, a significant gap remains in effectively deploying them in the highly unstructured and uncertain 3D physical world~\cite{chen2024spatialvlm, gholami2025spatial, qu2025spatialvla,yuan2025eoc,yu2025inst3d}.

\begin{wrapfigure}{r}{0.58\textwidth} 
  \centering
  \includegraphics[width=\linewidth]{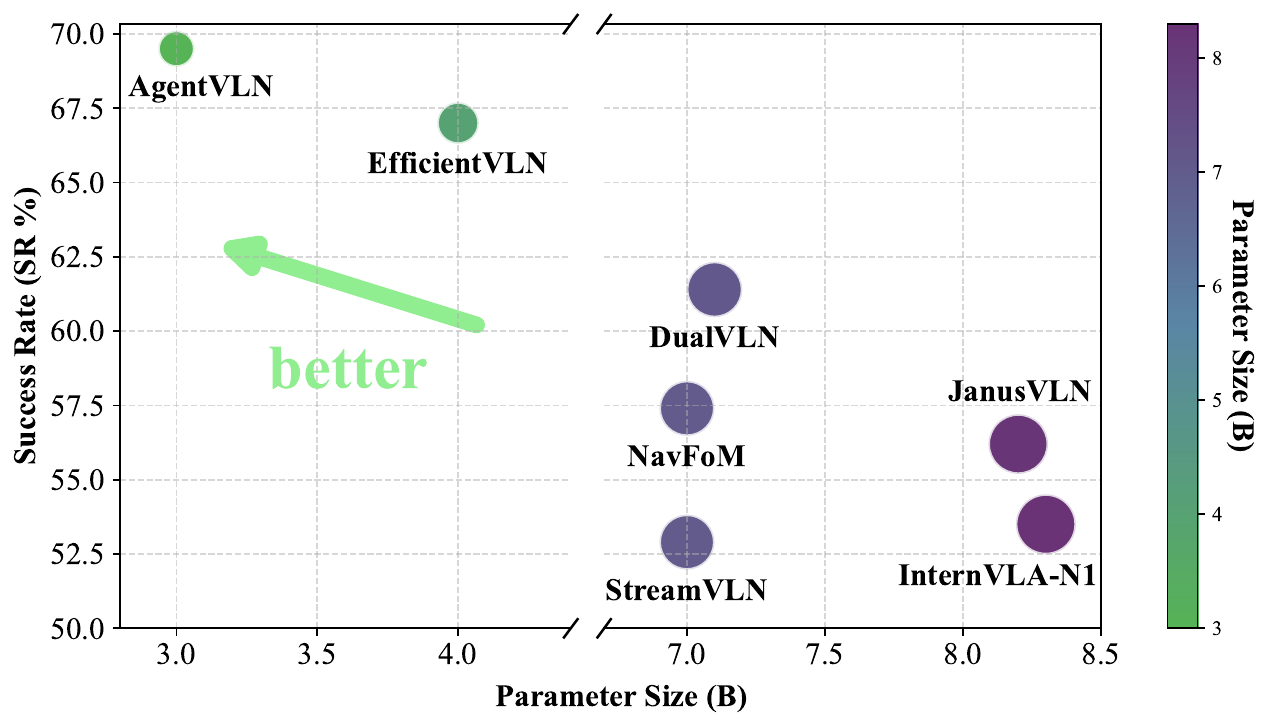}
  \vspace{-6mm}
  \caption{Performance comparison 
  on the Val-Unseen split of RxR-CE dataset~\cite{rxr}. AgentVLN outperforms existing state-of-the-art models while utilizing a substantially smaller parameter footprint. Furthermore, owing to its lightweight architecture, AgentVLN supports real-time local inference on Jetson embedded edge boards, eliminating the reliance on remote cloud deployment.}
  \label{fig:compare_param}
  \vspace{-17pt} 
\end{wrapfigure}

Current VLN systems can be primarily categorized into two paradigms: single-system and dual-system architectures. 
Despite their progress, both  still face critical challenges.
Single-system approaches~\cite{streamvln, janusvln, efficientvln} seek to implicitly map visual and linguistic features to action chunks using massive amounts of video-trajectory data. This black-box mapping 
struggles to learn 
the underlying geometric structure of navigation, limiting its cross-environment generalization.
\\Dual-system methods~\cite{internvla-n1, dualvln} typically introduce diffusion-based modules as trajectory generators while employing VLMs as low-frequency global planners. However, a pronounced cross-space disconnect frequently emerges between the generated 3D trajectories and the 2D visual representations of the VLMs. This stems from the fact that pre-trained VLMs inherently lack the capacity to reason about 3D geometry and metric scale. 
Moreover, both paradigms struggle with spatial scale ambiguity.
Monocular RGB observations lack reliable depth cues, making it hard for 
VLMs to localize targets involving spatial prepositions. While some methods~\cite{efficientvln, janusvln, internvla-n1} inject spatial information through additional depth-estimation modules~\cite{vggt, depthanything3}, this typically increases inference latency and context length, hindering efficient deployment.

To address the aforementioned issues, we propose \textbf{AgentVLN}, a novel and highly efficient embodied navigation framework. We reformulate the VLN system under a ``VLM-as-Brain'' paradigm that decouples high-level cognitive reasoning from low-level planning execution, and model the task as a Partially Observable Semi-Markov Decision Process (POSMDP). In AgentVLN, the VLM acts as the central controller and makes decisions based solely on the current multimodal context, performing global path navigation and local target localization.
By alternately invoking perception-level skills and cross-timestep planning-level skills, the system seamlessly scales from short-horizon exploration and error correction to long-horizon navigation, enabling robust generalization.

To tackle the multi-level representation inconsistency in dual-system VLN, we propose a cross-space representation mapping mechanism. In contrast to prior approaches that forcefully align multimodal features in an implicit high-dimensional latent space, we explicitly project 3D waypoints calculated by the perception layer onto the 
2D image plane via perspective projection, producing pixel-aligned visual prompts that are directly consumable by the VLM.
This design closes the semantic gap between 2D visual representations and 3D physical structure, significantly improving generalization in long-horizon navigation.
We further equip AgentVLN with a context-conditioned self-correction mechanism.
When the system encounters occlusions, blind spots, or trajectory deviations,
it generates fine-grained exploratory actions to actively recover, mitigating error accumulation and improving robustness in complex, unseen environments.

Furthermore, to mitigate spatial scale ambiguity during  local target localization, we introduce a Query-Driven Perceptual Chain-of-Thought (QD-PCoT) to
enable a VLN system to actively query for geometric cues on demand.
When confronted with spatial ambiguity, the agent actively formulates natural-language queries and invokes perception-layer skills. 
By integrating multi-round interactive feedback into an explicit chain-of-thought deduction, the agent ultimately infers instruction-consistent target pixel coordinates, which are then passed to planning-level skills to execute navigation.
Finally, we build AgentVLN-Instruct, a large-scale instruction-tuning dataset that tightly aligns high-level instructions with low-level skill invocations. It features a dynamic stage-routing mechanism conditioned on target visibility, explicitly mirroring the human wayfinding process: navigate coarsely first, localize precisely second. 

As shown in Fig.~\ref{fig:compare_param}, AgentVLN consistently outperforms prior state-of-the-art methods on the VLN-CE benchmark while using a substantially smaller parameter budget, delivering a superior accuracy–efficiency trade-off. Benefiting from its lightweight design, AgentVLN enables real-time, on-device inference on Jetson embedded edge platforms, removing the need for remote cloud deployment and the associated communication and decoding latency. Moreover, real-world evaluations further validate its robustness in long-horizon planning and efficient execution across diverse scenarios.

In summary, our key contributions are threefold:
\begin{itemize}
    \item We propose AgentVLN under a ``VLM-as-Brain'' paradigm, which decomposing long-horizon  navigation into high-level semantic reasoning and skill scheduling over a plug-and-play library, yielding strong generalization in unseen, complex environments.
    \item We introduce a cross-space representation mapping mechanism that converts the environment’s implicit 3D geometric topology into explicit, pixel-aligned visual prompts, and couple it with context-aware fine-grained self-correction and active exploration to enhance robustness.
    \item 
    We present a novel Query-Driven Perceptual Chain-of-Thought (QD-PCoT), enabling metacognitive target localization by actively querying for missing spatial cues during instruction grounding.
\end{itemize}

\section{Related Work}
\label{sec:2}
\subsection{Vision-and-Language Navigation}
VLN requires an embodied agent to follow natural-language instructions and precisely reach  a target destination using only egocentric visual observations. This complex task relies heavily on cross-modal semantic alignment and long-horizon spatial reasoning. Early approaches primarily utilize Seq2Seq-style architectures~\cite{seq2seq21, seq2seq22, seq2seq23} for direct state-to-action sequence predictions.  Subsequent works, such as NavGPT~\cite{navgpt} and InstructNav~\cite{instructnav}, integrate Large Language Models (LLMs) to handle high-level instruction parsing and global path planning. However, these non-end-to-end pipelines inevitably lose fine-grained geometric and   spatial cues 
during modality conversion. 
To address this gap, NaVid~\cite{zhang2024navid} introduces an end-to-end action prediction paradigm based on VLMs, while it remains constrained by context-length limits 
in long-horizon tasks. More recent methods, including StreamVLN~\cite{streamvln} and JanusVLN~\cite{janusvln}, alleviate this token explosion issue via sliding-window processing and slow-fast memory mechanisms. 
InternVLA-N1~\cite{internvla-n1} and DualVLN~\cite{dualvln} draw inspiration from human cognition and explore dual-system designs that decouple high-level semantic reasoning from low-level action execution, improving trajectory quality and navigation efficiency.
Despite these advances, practical deployment remains challenging. Current VLN systems~\cite{streamvln, janusvln, internvla-n1, cheng2024navila, zhang2024uninavid} typically operate at a large parameter scale (often $\ge$7B).
Furthermore, auxiliary depth  modules~\cite{efficientvln, janusvln} introduces substantial computational and storage overhead. 
Consequently, these methods are often deployed on remote cloud servers, where high-bandwidth data transmission and remote autoregressive generation make it difficult to satisfy the real-time control demands of physical agents in complex, unstructured environments.

\subsection{Vision-Language Model}
In recent years, VLMs have achieved great advancements in architecture designs and multimodal representation capabilities~\cite{llava,li2025eagle,li2025eagle,chen2025eagle,li2025tokenpacker,yuan2024osprey,wang2025videoitg,yuan2025videorefer,yuan2025pixelrefer}.  As a pioneering representative in this domain, the LLaVA series~\cite{llava, improvedllava, liu2024llavanext,li2024llava-one-vision} establish a widely adopted recipe for extending LLMs to visual perception tasks via visual instruction tuning. Subsequent variants further improve 
fine-grained visual understanding~\cite{liu2024llavanext,li2024llava-one-vision}, long-context video comprehension, and precise spatial grounding by introducing dynamic high-resolution processing alongside more comprehensive training corpora. 
More recently, VLMs are undergoing a paradigm progressively evolving from passive perception and understanding tools into active, decision-making agents~\cite{qwen2_vl, qwen3, qwen25_vl}. For instance, Qwen3-VL~\cite{qwen3, qwen3_vl} and Qwen3.5~\cite{qwen35} integrate ``thinking with images'' and function-calling mechanisms, facilitating tool-augmented self-reflection and sophisticated multi-step reasoning. Nevertheless, a glaring limitation persists: current multimodal agent methodologies are almost exclusively constrained to highly structured software environments~\cite{osworld, androidworld, screenspot} and inherently fail to generalize to the dynamic, unstructured complexities of real-world physical scenarios.

\section{AgentVLN}
\label{sec:method}

% \subsection{VLM-as-Brain}
\subsection{Overview}
\label{sec:31}
\begin{figure*}[]
  \centering
   \includegraphics[width=1.0\linewidth]{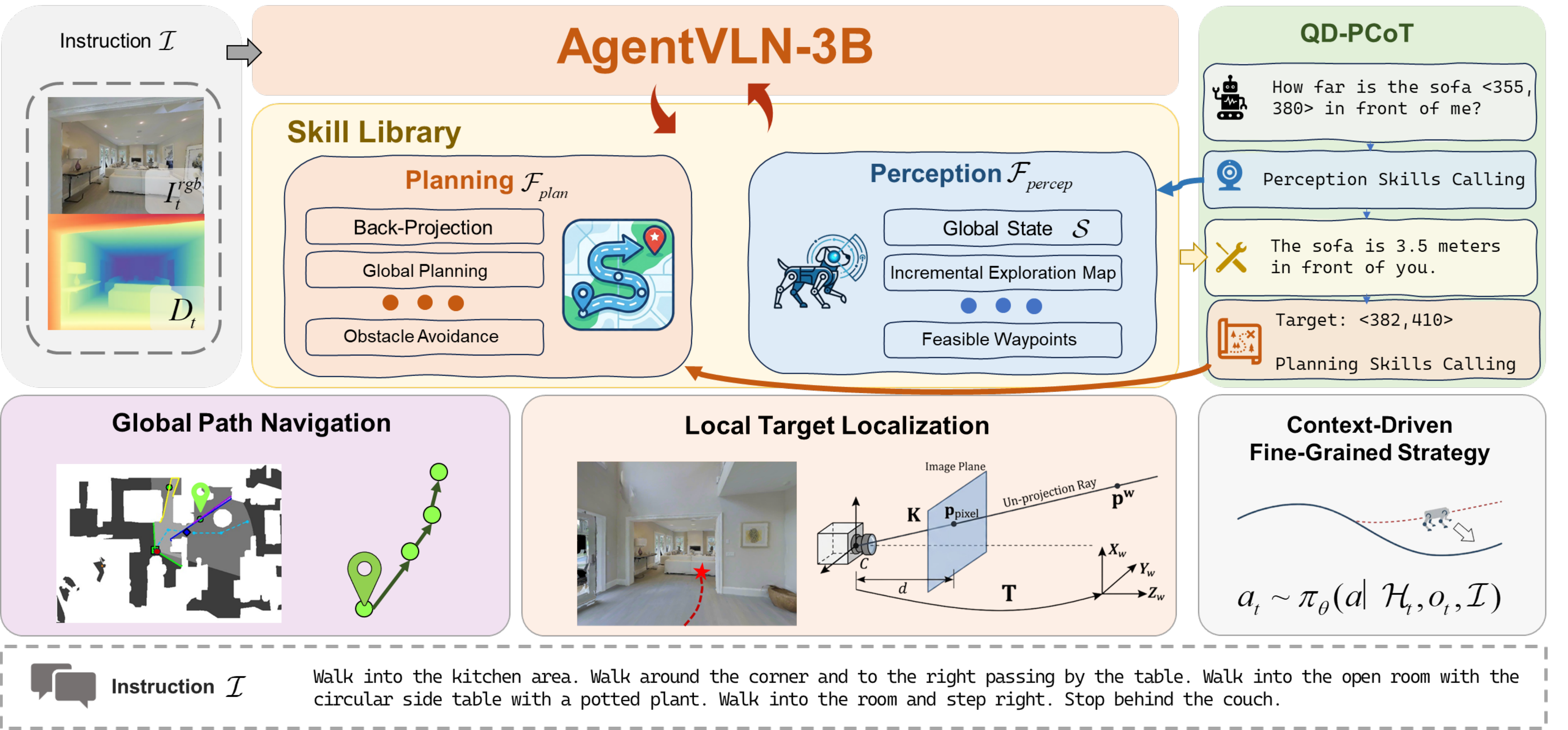}
   \caption{Overview of the AgentVLN framework. AgentVLN employs a VLM-as-Brain paradigm, decomposing long-horizon navigation into modular skill executions. 
   Additionally, a context-driven fine-grained strategy and QD-PCoT mitigate localization errors and scale ambiguities, ensuring precise 3D target grounding.
   }
   \label{fig:framework}
   \vspace{-0.3cm}
\end{figure*}

In this section, we present AgentVLN, a novel and efficient embodied navigation framework. Unlike prevailing end-to-end approaches that require massive video datasets for pre-training to implicitly learn spatial geometry into policies, AgentVLN adopts a \textbf{VLM-as-Brain} embodied paradigm. This paradigm explicitly decouples high-level cognitive reasoning from low-level skill execution. Specifically, we leverage a VLM as the central controller to interpret complex natural-language instructions and decompose long-horizon navigation into a structured sequence of skill invocations.
As a general-purpose embodied agent, AgentVLN can generalize 
to various downstream tasks, including instruction-guided navigation, autonomous exploration, and semantic mapping. 
Moreover, AgentVLN abstracts the underlying motion mechanics into a highly modular and plug-and-play skill library.  As a result, 
adapting to novel environments or tasks 
can be achieved by updating or expanding the skill library, without  retraining the multi-billion-parameter VLM.

\subsection{Problem Definition}
We formalize AgentVLN's 
decision-making process as a Partially Observable Semi-Markov Decision Process (POSMDP) jointly conditioned on natural language and temporal memory, defined as $\mathcal{M} = \langle \mathcal{S}, \mathcal{O}, \mathcal{F}, \mathcal{T}, \mathcal{I}, \mathcal{H} \rangle$. The state $s \in \mathcal{S}$ encodes the agent's global position and pose, while the observation $o_t \in \mathcal{O}$ comprises  
% the set of 
egocentric data streams acquired from onboard sensors at time step $t$. 
The agent's action space is no longer defined by low-level continuous velocity commands or discrete directional movements (\textit{e.g.}, forward/left/right), but rather by a set of skill functions $\mathcal{F} = \{f_1, f_2, \dots, f_K\}$. 
The multi-step transition dynamics $\mathcal{T}$ describe the state evolution laws over a variable time steps $\tau$ 
% following the execution of a skill, 
induced by executing a skill, denoted as $P(s_{t+\tau}, \tau \mid s_t, f)$. Concretely, at each decision step $t$, the agent generates a high-level control policy or a structured calling instruction to activate a skill based on the current history context $\mathcal{H}_t$, visual observations $o_t$, and user instructions $\mathcal{I}$:
\begin{small}
\begin{equation}
    c_k \sim \pi_{\theta}(f \mid \mathcal{H}_{t_k}, o_{t_k}, \mathcal{I}), \quad f \in \mathcal{F},
\end{equation}
\end{small}
where $\pi_{\theta}$ is 
the policy model parameterized by VLM.

Motivated by the human cognitive pattern of  ``perceive before act'', we further decompose the skill set into perception-level skills $\mathcal{F}_{percep}(\tau=0)$ and planning-level skills $\mathcal{F}_{plan}(\tau > 0)$, with $\mathcal{F} = \mathcal{F}_{percep} \cup \mathcal{F}_{plan}$. 
Perception-level skills extract geometric or semantic features from the environment and return augmented observations $\Delta o$, which are integrated into the context to reduce local ambiguity and inform subsequent decisions. In contrast, planning-level skills perform closed-loop physical execution: given a skill policy $\pi_{f}$, they produce a sequence of low-level actions $a_t$ over $\tau$ time steps, following the multi-step transition distribution: 
\begin{scriptsize}
\begin{equation}
    P(s_{t_k+\tau}, \tau \mid s_{t_k}, f_{c_k}) = \sum_{a_1 \dots a_\tau} \left( \prod_{i=0}^{\tau-1} P(s_{t_k+i+1} \mid s_{t_k+i}, a_{t_k+i}) \pi_{f_{c_k}}(a_{t_k+i}) \right) \beta_{f_{c_k}}(s_{t_k+\tau}).
\end{equation}
\end{scriptsize}Here, $\pi_{f_{c_k}}(a_{t_k+i})$ is the control signal produced by the navigation skill at step $i$, $P(s_{t_k+i+1} \mid s_{t_k+i}, a_{t_k+i})$ denotes the local state transition under action $a_{t_k+i}$, 
and $\beta_{f_{c_k}}(s_{t_k+\tau}) \in [0, 1]$ is the skill termination condition that determines whether the VLM needs to initiate a new decision cycle. If $\beta = 0$, the current planning skill continues; otherwise, the VLM updates the state context and proceeds to the next semi-Markov decision-making cycle.

In VLN, 
the VLM is optimized to precisely translate high-dimensional, complex semantic intentions with spatiotemporal motion trajectories in the physical space. 
In the initial stages of an episode, the VLM tends to alternately invoke perception and planning skills. First, it calls upon $\mathcal{F}_{percep}$ given $\mathcal{O}$ to construct a local geometric map.  % \textcolor{red}{geometric/semantic} map. 
Subsequently, incorporating the spatial prepositions within $\mathcal{I}$, the VLM triggers a planning skill to execute a $\tau$-step movement. When the visual features in the observation $o_t$ highly match the destination description in the navigation instruction $\mathcal{I}_{nav}$, the decision-making process automatically switches from global path navigation to local target localization, guiding the robot to accomplish precise docking.

\subsection{Global Path Navigation}
\label{sec:32}
To address 3D trajectory failures arising from multi-level representation inconsistencies, typically of popular dual-system architectures, AgentVLN introduces a novel cross-space representation mapping mechanism. 
During navigation, the VLM first invokes the perception-level skill $\mathcal{F}_{percep}$ to update the agent's global state $\mathcal{S}$ and construct an incremental exploration map.
At each time step $t$, the agent receives multimodal observations $o_t = \{I_t^{rgb}, D_t\}$ together with its pose in the world coordinate system $T_t = [R_t \mid \mathbf{t}_t] \in SE(3)$, where $R_t \in SO(3)$ is the rotation matrix and $\mathbf{t}_t = [x_t, y_t, h_c]^T$ is the translation vector with   camera height $h_c$. Given the camera intrinsic matrix $K \in \mathbb{R}^{3 \times 3}$, for any valid pixel $\mathbf{p}_{img} = [u, v, 1]^T$ with depth $d$ in 
$D_t$, we back-project
it into a 3D point $\mathbf{P}^w$ in the world coordinate system via 
\begin{equation}
    \mathbf{P}^w = R_t (d \cdot K^{-1} \mathbf{p}_{img}) + \mathbf{t}_t.
\end{equation}

As the agent moves, these back-projected points are incrementally updated into a global occupancy grid map,  enabling planning-level skills to generate a series of 3D waypoints. 
To enable the VLM to make navigation decisions directly within the visual space, let a globally feasible path point in the world coordinate system be denoted as $\mathbf{P}_{path}^w = [X_{path}, Y_{path}, 0]^T$. Through the inverse transformation of the camera pose $T_t$ and perspective projection, we calculate the pixel coordinates $\mathbf{p}_{path}^{img} = [u_{path}, v_{path}, 1]^T$ of this 3D path point on the current observation:
\begin{equation}
    s \cdot \mathbf{p}_{path}^{img} = K R_t^{-1} (\mathbf{P}_{path}^w - \mathbf{t}_t),
\end{equation}
where $s$ is the depth scale factor of the mapped point in the camera coordinate system. By converting the  environment's implicit 3D topology into explicit, pixel-aligned visual prompts, 
the VLM only needs to select the matching path in the pixel space based on the semantics of $\mathcal{I}$. 
It then invokes the planning-level skills $\mathcal{F}_{plan}$ to restore this selection to a 3D waypoint for navigation control. 
This mechanism not only resolves the VLM's inherent deficit in spatial perception but also significantly improves generalization in long-horizon navigation.

Furthermore, to mitigate the issues, like view occlusions, instruction ambiguity, and accumulated localization errors in complex environments, we present  a context-driven, fine-grained closed-loop self-correction and active exploration mechanism. When the current observation $o_t$ contains no globally feasible path projection that is semantically relevant to $\mathcal{I}$, 
the agent performs commonsense reasoning and local geometric inference conditioned on  $\mathcal{H}_t$ and  $\mathcal{I}$, and outputs fine-grained actions:
\begin{equation}
    a_t \sim \pi_{\theta}(a \mid \mathcal{H}_t, o_t, \mathcal{I}), \quad a \in \{\texttt{Forward, Left, Right}\}.
\end{equation}
Importantly, in blind spots without feasible global path guidance, the agent is not forced to execute long-distance blind displacements. 
Instead, this context-based fine-tuning enables the agent to autonomously look around and to promptly correct its course when deviating from the intended route. Once a valid global path mapping is successfully captured, AgentVLN resumes alternating between perception- and planning-level skills for efficient long-horizon execution. 
This dynamic switching between macro-skill invocation and context-driven corrective fine-tuning endows AgentVLN with both high efficiency and strong self-recovery, improving robustness under trajectory deviations.

\subsection{Local Target Localization}
\label{sec:33}
As navigation proceeds, once the agent’s observation $o_t$ contains multimodal evidence that strongly matches the destination description in the instruction $\mathcal{I}$, 
AgentVLN smoothly transitions from exploration-based global path planning  to local target localization. 
In this stage, the VLM locks onto the destination target coordinates in the egocentric 2D view, and achieves precise docking through spatial geometric remapping.

However, in real-world unstructured environments, instructions frequently exhibit spatial ambiguity. Under such ambiguity, relying solely on monocular RGB makes it difficult for the VLM to accurately predict the target's pixel coordinates, frequently leading to failures due to scale uncertainty and depth-perspective ambiguity. 
To address this key limitation, we propose a novel Query-Driven Perceptual Chain-of-Thought (QD-PCoT), endowing the agent with the metacognitive ability to actively query missing information.

Specially, when the model detects spatial ambiguity, it does not blindly output coordinates. Instead, it generates an intermediate natural-language query \textit{(e.g., ``How many meters away is the chair in front of me?'')} and triggers a specific perception-level skill $\mathcal{F}_{percep}$. The returned geometric/semantic information  is then fed back into the context as an incremental textual prompt. 
Guided by this multi-step reasoning process, 
the model then outputs precise destination pixel coordinates $\mathbf{p}_{target}^{img} = [u_{target}, v_{target}, 1]^T$ that strictly align with the instructional intent. To project the pixel coordinates into 3D coordinate system, we extract its depth $d_{target} = D_t(u_{target}, v_{target})$ from the depth map $D_t$ and back-project it into the local camera coordinate system:
\begin{equation}
    \mathbf{P}_{target}^c = d_{target} \cdot K^{-1} \mathbf{p}_{target}^{img} = \begin{bmatrix} \frac{(u_{target} - c_x) \cdot d_{target}}{f_x} \\ \frac{(v_{target} - c_y) \cdot d_{target}}{f_y} \\ d_{target} \end{bmatrix},
\end{equation}
where $f_x$, $f_y$ are focal lengths, and $c_x$, $c_y$ represent the principal point. Given the current pose $T_t$, we transform the 3D point to the global coordinate system via a rigid-body transformation:
\begin{equation}
    \mathbf{P}_{target}^w = R_t \mathbf{P}_{target}^c + \mathbf{t}_t.
\end{equation}

% Through this process, 
In this way, the VLM's high-level semantic selection in the 2D visual plane is seamlessly translated into 3D target coordinates $\mathbf{P}_{target}^w$ in physical space, after which planning-level skills are invoked to execute accurate goal-reaching and docking.

\subsection{AgentVLN-Instruct Dataset}
To address the instruction-skill alignment gap, we construct AgentVLN-Instruct, a large-scale dataset based on Habitat. Deviating from singular control policies, it introduces a dynamic stage-routing mechanism conditioned on target visibility in $I_t^{rgb}$, emulating coarse-to-fine human wayfinding. In global path planning, the model invokes perception skills $\mathcal{F}_{percep}$ to select expert-aligned waypoints. To foster robust self-localization and closed-loop error correction, we inject trajectory noise and utilize fine-grained action sequences as fallbacks when visual waypoints are absent. In local target localization, a multi-round perceptual Chain-of-Thought (CoT) simulates active reasoning to predict precise target pixel coordinates. Ultimately, by activating pre-trained VLM commonsense via explicit CoT and modular skills, this natural-language-guided paradigm effectively bridges 2D-3D modalities. It significantly enhances both localization success and interpretability in ambiguous environments, establishing a robust framework for embodied agents. Please refer to the \textit{\textit{supplemental material}} for more details.

\begin{table*}[tb]
\centering
\caption{Comparison results with SOTA methods on the Val-Unseen dataset for R2R-CE. Our method outperforms other approaches on the same benchmarks, even having fewer parameters.}
\begin{tabular}{lcccccccc}
\hline
\multirow{2}{*}{Method} & \multicolumn{4}{c}{Observation}   & \multicolumn{4}{c}{R2R-CE} \\ \cline{2-9}
             & S.RGB  & Pano.  & Depth  & Odo.   & NE ↓ & OS ↑ & SR ↑ & SPL ↑ \\ \hline
HPN+DN ~\cite{krantz2021waypoint}  &   & \checkmark & \checkmark & \checkmark & 6.31 & 40.0 & 36.0 & 34.0  \\
VLN BERT ~\cite{hong2022bridging}  &   & \checkmark & \checkmark & \checkmark & 5.74 & 53.0 & 44.0 & 39.0  \\
CMA ~\cite{hong2022bridging}       &   & \checkmark & \checkmark & \checkmark & 6.20 & 52.0 & 41.0 & 36.0  \\
Reborn ~\cite{an20221st}           &   & \checkmark & \checkmark & \checkmark & 5.40 & 57.0 & 50.0 & 46.0  \\
Ego$^{2}$-Map ~\cite{hong2023learning} &   & \checkmark & \checkmark & \checkmark & 5.54 & 56.0 & 47.0 & 41.0  \\
DreamWalker ~\cite{wang2023dreamwalker} &  & \checkmark & \checkmark & \checkmark & 5.53 & 59.0 & 49.0 & 44.0  \\
ETPNav ~\cite{an2023etpnav}        &   & \checkmark & \checkmark & \checkmark & 4.71 & 65.0 & 57.0 & 49.0  \\ \hline
Seq2Seq ~\cite{krantz2020beyond}   & \checkmark &   & \checkmark &   & 7.77 & 37.0 & 25.0 & 22.0  \\
RGB-CMA ~\cite{krantz2020beyond}   & \checkmark &   &   &   & 9.55 & 10.0 & 5.0  & 4.0   \\
AG-CMTP ~\cite{chen2021topological}&   & \checkmark & \checkmark &   & 7.90 & 39.0 & 23.0 & 19.0  \\
R2R-CMTP ~\cite{chen2021topological}&  & \checkmark & \checkmark &   & 7.90 & 38.0 & 26.0 & 22.0  \\
LAW ~\cite{raychaudhuri2021law}    & \checkmark &   & \checkmark & \checkmark & 6.83 & 44.0 & 35.0 & 31.0  \\
CM2 ~\cite{georgakis2022cross}     & \checkmark &   & \checkmark &   & 7.02 & 41.0 & 34.0 & 27.0  \\
WS-MGMap ~\cite{chen2022weakly}    & \checkmark &   & \checkmark &   & 6.28 & 47.0 & 38.0 & 34.0  \\
ETPNav+FF ~\cite{wang2024sim}      & \checkmark &   & \checkmark &   & 5.95 & 55.8 & 44.9 & 30.4  \\
AO-Planner ~\cite{chen2024affordances} &   & \checkmark & \checkmark &   & 5.55 & 59.0 & 47.0 & 33.0  \\ \hline
NaVid-7B ~\cite{zhang2024navid}       & \checkmark &   &   &   & 5.47 & 49.0 & 37.0 & 35.0  \\
Uni-NaVid-7B ~\cite{zhang2024uninavid}& \checkmark &   &   &   & 5.58 & 53.3 & 47.0 & 42.7  \\
NaVILA-7B ~\cite{cheng2024navila}     & \checkmark &   &   &   & 5.22 & 62.5 & 54.0 & 49.0  \\
StreamVLN-7B ~\cite{streamvln}        & \checkmark &   &   &   & 5.10 & 64.0 & 55.7 & 50.9  \\
% DecoVLN-7B~\cite{cvpr2026decovln}  & \checkmark &   &   &   &   \\
NavFoM-7B ~\cite{zhang2025embodied}   & \checkmark &   &   &   & 5.01 & 64.9 & 56.2 & 51.2  \\
DecoVLN-7B~\cite{cvpr2026decovln}     & \checkmark &   &   &   & 5.01 & 63.5 & 56.3 & 50.5 \\
JanusVLN-8.2B ~\cite{zeng2025janusvln}& \checkmark &   &   &   & 4.78 & 65.2 & 60.5 & 56.8  \\
EfficientVLN-4B~\cite{efficientvln}   & \checkmark &   &   &   & 4.18 & \textbf{73.7} & 64.2 & 55.9  \\
DualVLN-7.1B~\cite{dualvln}                & \checkmark &   &   &   & 4.05 & 70.7 & 64.3 & 58.5  \\
StreamVLN-7B ~\cite{streamvln}        & \checkmark &   & \checkmark &   & 4.98 & 64.2 & 56.9 & 51.9  \\
InternVLA-N1-8.3B~\cite{internvla-n1}    & \checkmark &   & \checkmark &   & 4.83 & 63.3 & 58.2 & 54.0  \\
\rowcolor[HTML]{EAFFFE} 
\textbf{AgentVLN-3B}                  & \checkmark &   & \checkmark &   & \textbf{3.88} & 73.5 & \textbf{67.2} & \textbf{64.7}  \\ \hline
\end{tabular}
\label{tab:r2r_results}
\vspace{-0.5cm}
\end{table*}

\section{Experiment}
\label{sec:exp}
\subsection{Datasets and Evaluation Metrics}
To evaluate the model under varying levels of navigation difficulty and semantic complexity, we conduct extensive experiments on the benchmark datasets R2R-CE~\cite{r2r} and RxR-CE~\cite{rxr}. However, existing VLN datasets only provide coarse instruction–waypoint mappings, which are insufficient to support the finetuning requirements of high-level skill invocation and QD-PCoT reasoning in the VLM-as-Brain architecture. 
To address this limitation, we construct a large-scale instruction tuning dataset, AgentVLN-Instruct, based on the Habitat simulator~\cite{habitat2021}. Unlike traditional trajectory collection approaches, this dataset incorporates four key components: target-visibility-driven dynamic routing, generalizable skill invocation, localization reasoning, and active question–answer interactions. 
To preserve the model’s general multimodal capabilities, we additionally include the LLaVA-Video-178K dataset~\cite{llava_video_178k} during training. For evaluation, we follow standard VLN benchmarking protocols and report commonly used navigation metrics, including Success Rate (SR), Oracle Success Rate (OS), Success weighted by Path Length (SPL), and Navigation Error (NE).

\subsection{Implementation Details}
We adopt Qwen2.5-VL-3B~\cite{qwen25_vl} as the brain of AgentVLN. During the instruction-tuning phase, we freeze the model's visual encoder and multimodal projection layer. The network is optimized using the AdamW~\cite{adamw} with a batch size of 128. For the learning rate schedule, we employ a cosine annealing strategy with a peak learning rate of $2 \times 10^{-5}$, incorporating a warmup ratio of 0.03 at the initial stage of training. The egocentric observations are captured at a resolution of $640 \times 480$ with a 110° field of view, and the temporal window for the multimodal historical context is set to 8 frames. All experiments were conducted on a computing cluster equipped with 32 NVIDIA A100 GPUs. Please refer to the \textit{\textit{supplemental material}} for more details.

\subsection{Navigation Experiments}

\begin{table*}[t]
\centering
\caption{Comparison with SOTA methods on RxR-CE Val-Unseen split.}
\begin{tabular}{lcccccccc}
\hline
Method      & S.RGB     & Pano. & Depth     & Odo.  & NE $\downarrow$ & SR $\uparrow$ & SPL $\uparrow$ & nDTW $\uparrow$ \\ \hline
VLN BERT~\cite{hong2022bridging}  &   & \checkmark & \checkmark & \checkmark & 8.98    & 27.0  & 22.6   & 46.7    \\
CMA~\cite{hong2022bridging}   &   & \checkmark & \checkmark & \checkmark & 8.76    & 26.5  & 22.1   & 47.0    \\
Reborn ~\cite{an20221st}  &   & \checkmark & \checkmark & \checkmark & 5.98    & 48.6  & 42.0   & 63.3    \\
ETPNav~\cite{an2023etpnav}    &   & \checkmark & \checkmark & \checkmark & 5.64    & 54.7  & 44.8   & 61.9    \\ \hline
Seq2Seq~\cite{krantz2020beyond}   & \checkmark &   & \checkmark &   & 12.10   & 13.9  & 11.9   & 30.8    \\
LAW~\cite{raychaudhuri2021law}    & \checkmark &   & \checkmark & \checkmark & 10.90   & 8.0   & 8.0    & 38.0    \\
ETPNav+FF~\cite{wang2024sim}  & \checkmark &   & \checkmark & \checkmark & 8.79    & 25.5  & 18.1   & -   \\
AO-Planner~\cite{chen2024affordances} &   & \checkmark & \checkmark &   & 7.06    & 43.3  & 30.5   & 50.1    \\ \hline
Uni-NaVid-7B~\cite{zhang2024uninavid}    & \checkmark &   &   &   & 6.24    & 48.7  & 40.9   & -   \\
NaVILA-7B~\cite{cheng2024navila} & \checkmark &   &   &   & 6.77    & 49.3  & 44.0   & 58.8    \\
NavFoM-7B~\cite{zhang2025embodied}  & \checkmark &   &   &   & 5.51    & 57.4  & 49.4   & 60.2    \\
JanusVLN-8.2B~\cite{janusvln}    & \checkmark &    &    &    & 6.06    & 56.2  & 47.5   & 62.1    \\
StreamVLN-7B~\cite{streamvln}    & \checkmark &   &   &   & 6.16    & 51.8  & 45.0   & 61.9    \\
% DecoVLN-7B~\cite{cvpr2026decovln}  & \checkmark &   &   &   &  \\ 
InternVLA-N1-7B~\cite{internnav} & \checkmark &   &   &   & 6.41    & 49.5  & 41.8   & 62.6    \\
DecoVLN-7B~\cite{cvpr2026decovln} & \checkmark &   &   &   & 5.73 & 54.2 & 46.3 & 63.5  \\
EfficientVLN-4B~\cite{efficientvln}    & \checkmark &    &    &    & \textbf{3.88}    & 67.0  & 54.3   & 68.4    \\
DualVLN-7.1B~\cite{dualvln}      & \checkmark &   &  &   &   4.58 & 61.4 & 51.8  & 70.0 \\
StreamVLN-7B~\cite{streamvln}    & \checkmark &    & \checkmark &    & 6.22    & 52.9  & 46.0   & 61.9    \\
InternVLA-N1-8.3B~\cite{internvla-n1}    & \checkmark &    & \checkmark &    & 5.91    & 53.5  & 46.1   & 65.3    \\
\rowcolor[HTML]{EAFFFE} 
\textbf{AgentVLN-3B}            & \checkmark &   & \checkmark &   & 3.92 & \textbf{69.5} & \textbf{61.3} & \textbf{74.6}  \\ \hline
\end{tabular}
\label{tab:rxr_results}
\vspace{-0.1cm}
\end{table*}

\begin{figure*}[]
  \centering
   \includegraphics[width=1.0\linewidth]{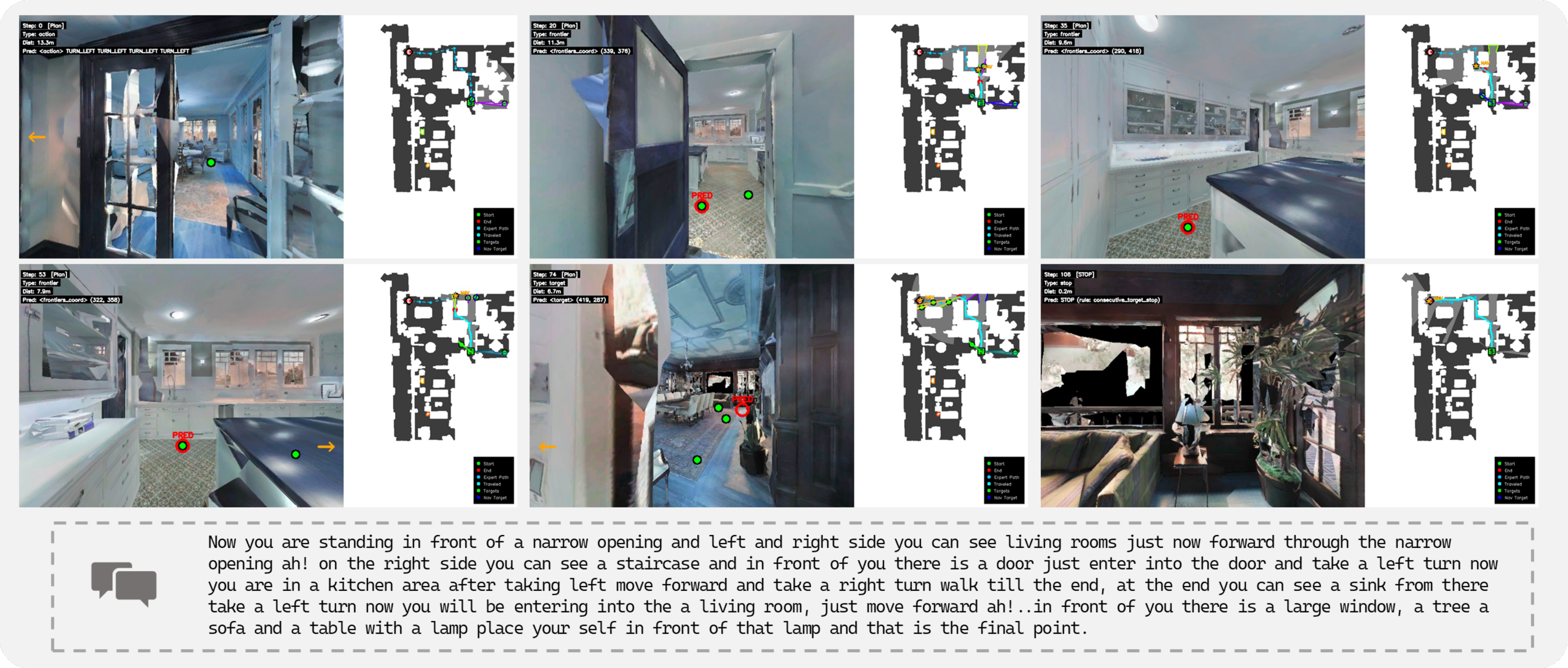}
   \caption{Visualization of AgentVLN's navigation.  Green points represent the visual prompts generated by the perception-level skills, whereas red circles denote the navigation waypoints selected or predicted by the model. Notably, when traversing narrow passages or confronting severe visual occlusions, the model seamlessly outputs fine-grained atomic actions for trajectory fine-tuning. This demonstrates its robust capability to achieve highly accurate, collision-free navigation relying exclusively on egocentric observations.}
   \label{fig:sim}
   \vspace{-7mm}
\end{figure*}

 \cref{tab:r2r_results} and~\cref{tab:rxr_results} present the performance of AgentVLN compared against SOTA methods on the Val-Unseen splits of the R2R-CE~\cite{r2r} and RxR-CE~\cite{rxr} benchmarks. The evaluated baselines encompass waypoint prediction models relying on multi-sensor inputs, as well as large vision-language navigation models. Operating with a remarkably compact parameter scale of merely 3B, AgentVLN outperforms the SOTA method of the same class, InternVLA, by substantial margins of 9.0\% SR and 10.7\% SPL on the R2R dataset, all while maintaining a significantly smaller parameter footprint. 
To compensate for the VLM's inherent deficit in 3D scale perception within the physical world, prior methods such as EfficientVLN introduce monocular depth estimation modules to compel the VLM to implicitly learn 3D geometric relationships. However, this approach not only incurs severe computational overhead but also heavily exacerbates the model's memory burden. In contrast, on the R2R and RxR datasets, AgentVLN achieves absolute improvements of 3.0 and 2.5 in SR over EfficientVLN, respectively. More notably, on the SPL metric which strictly reflects the optimality of the navigation trajectory and overall decision-making efficiency. Our method yields impressive improvements of 8.8\% and 7.0\%. These results explicitly validate that the VLM-as-Brain paradigm, when integrated with context-driven fine-grained self-correction, effectively elevates navigation efficiency. We display the visualization results in the ~\cref{fig:sim}.

In conclusion, the experimental results substantiate that AgentVLN significantly bolsters navigation performance with absolutely zero additional parameter overhead, thereby establishing a novel and highly effective paradigm for next-generation, lightweight, and high-precision embodied navigation frameworks.

\subsection{Real-World Deployment}
\begin{figure*}[t]
  \centering
   \includegraphics[width=1.0\linewidth]{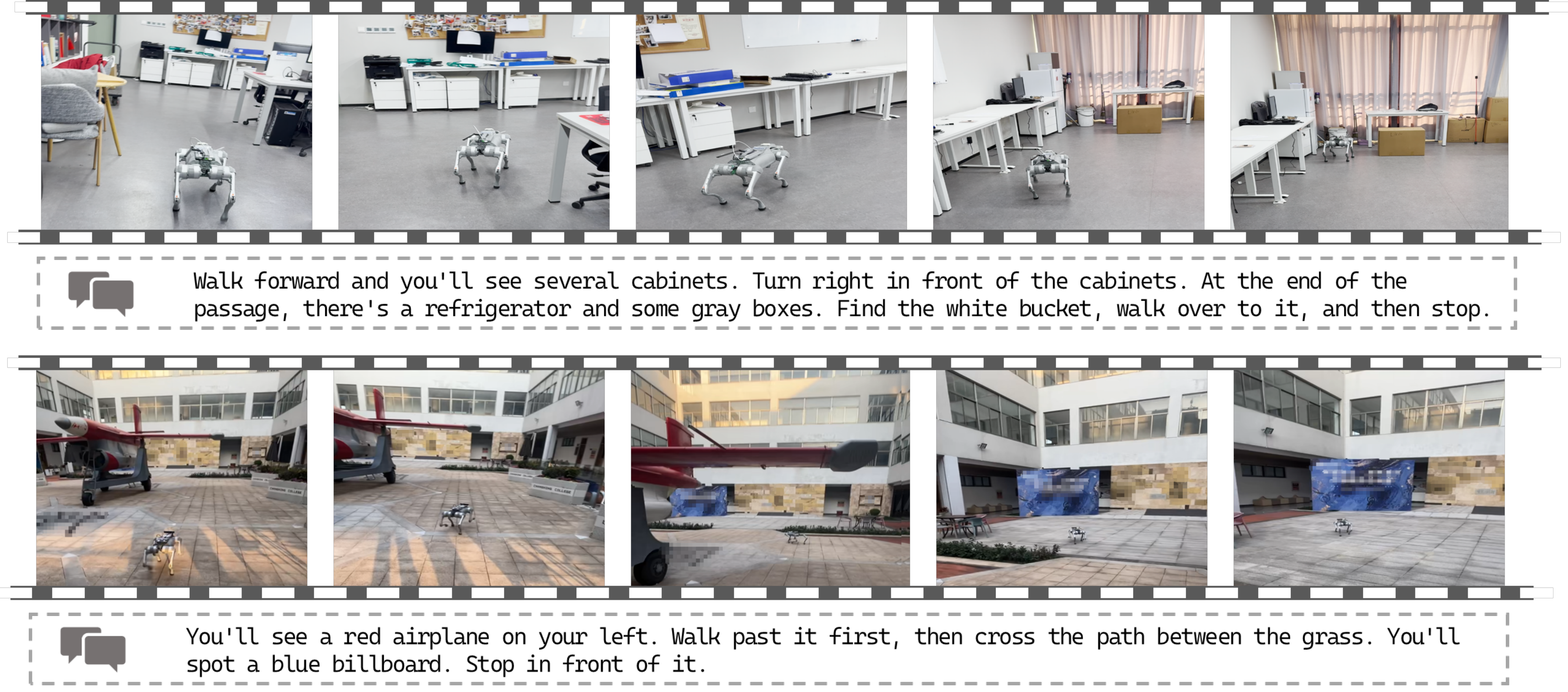}
   \caption{Navigation results in real-world indoor and outdoor environments.  Experimental results demonstrate that, regardless of whether the agent is navigating through complex, confined indoor spaces or outdoor scenarios with challenging illumination conditions, the proposed model consistently and accurately comprehends natural language instructions, enabling it to rapidly plan and execute precise navigation trajectories.}
   \label{fig:real_world}
   \vspace{-0.8cm}
\end{figure*}

% Figure: real-world
Existing VLN models, burdened by massive parameter counts and a reliance on auxiliary 3D vision encoders, typically necessitate offloading data streams to cloud servers for asynchronous inference. The compounding of autoregressive generation latency from large models and network communication delays renders these models inadequate for the high-frequency control demands of real-world scenarios. In stark contrast, AgentVLN adopts Qwen2.5-VL-3B~\cite{qwen25_vl} as its base model and eschews any 3D feature fusion modules, thereby enabling fully localized execution on Jetson edge computing platforms. As shown in ~\cref{fig:real_world}, and experimental results demonstrate that AgentVLN exhibits outstanding navigation performance in both indoor and outdoor environments.

We employ the Unitree Go2 quadruped robot as our physical platform, equipped with an Intel RealSense D455 camera to acquire the egocentric observation $\mathbf{O}$.  The perception-layer skill library incorporates the RTAB-Map~\cite{rtab_map} algorithm alongside a suite of camera processing utilities. Upon the VLM triggering a perception skill, RTAB-Map executes real-time local SLAM and incrementally updates the 3D occupancy grid map.  Concurrently, inverse perspective projection is utilized to supply the VLM with pixel-level visual prompts. Once the VLM outputs the target pixel coordinates, these coordinates are instantaneously remapped into 3D target waypoints within the global coordinate system. Subsequently, the planning-level skills encompassing path generation algorithms and robot control software packages are invoked to generate smooth, obstacle-avoiding trajectories, which are then translated into low-level control signals. Notably, this highly modular methodology is readily generalizable to other sensor modalities such as LiDAR, adapting to new sensors merely requires substituting the relevant perception-layer skills within the library, entirely circumventing the need to retrain the multi-billion-parameter foundation model. Please refer to the \textit{supplemental material} for detailed hardware specifications and  configurations.
% algorithm

\subsection{Ablation Studies}
\subsubsection{Incremental Ablation Study.}
To comprehensively dissect the individual contributions of each module within the AgentVLN architecture to the overall system navigation performance, we conducted systematic ablation studies on the R2R validation set. As presented in the ~\cref{tab:ablation_study}, under an end-to-end paradigm devoid of explicit geometric guidance, lightweight VLMs struggle profoundly to implicitly learn complex 3D spatial geometric relationships and physical reachability solely from the appearance features of 2D images. Consequently, the baseline model achieves 38.6\% SR merely, exposing severe deficiencies in the VLM's spatial perception capabilities. 
Building upon the baseline, the introduction of the cross-space representation mapping mechanism and the establishment of the VLM-as-Brain high-level scheduling paradigm yield substantial improvements: SR and OS surge by absolute margins of 21.1\% and 16.6\%, respectively, while NE decreases significantly to 4.67m. This demonstrates that alleviating the VLM's cognitive load associated with processing low-level action spaces and implicit spatial reasoning, thereby allowing it to focus exclusively on high-level semantic-visual matching and skill scheduling can significantly enhance model performance. 
Furthermore, the incorporation of the context-driven fine-grained strategy propels the SR steadily to 65.6\%, while the NE is further compressed to 3.90m. The fundamental driver behind this performance leap lies in its effective mitigation of the irreversible error accumulation problem typical of long-horizon tasks.
Finally, by integrating the QD-PCoT mechanism during the local target localization stage, the SR and SPL are further elevated to 67.2\% and 64.7\%, respectively. This indicates that in complex real-world scenarios, directly regressing target coordinates from 2D images frequently suffers from severe deviations induced by scale hallucinations. The QD-PCoT mechanism ensures that when confronted with complex environments, the VLM no longer resorts to blind, black-box predictions. Instead, it actively invokes perception-layer skills to acquire informational feedback, thereby elegantly eliminating the depth ambiguity of 2D vision without incurring any additional parameter overhead.

\begin{table}[t]
\centering
\small
\caption{Ablation study of different modules, where CDFG denotes context-driven fine-grained strategy.}
% \begin{tabular}{ccccccc}
\begin{tabular*}{\textwidth}{@{\extracolsep{\fill}}ccccccc}
\hline
VLM-as-Brain & CDFG & QD-PCoT & NE ↓ & OS↑   & SR↑   & SPL↑  \\ \hline
             &      &         & 6.53 & 48.50 & 38.60 & 35.10 \\
\checkmark            &      &         & 4.67 & 65.10 & 59.70 & 55.60 \\
\checkmark            & \checkmark    &         & 3.90 & 72.10 & 65.60 & 63.40 \\
\checkmark            & \checkmark    & \checkmark       & \textbf{3.88} & \textbf{73.50} & \textbf{67.20} & \textbf{64.70}  \\ \hline
\end{tabular*}
\vspace{-0.5cm}
\label{tab:ablation_study}
\end{table}

\subsubsection{Temporal Context Length.}
\begin{wrapfigure}{r}{0.6\textwidth} 
  \vspace{-2pt} 
  \centering
  \includegraphics[width=\linewidth]{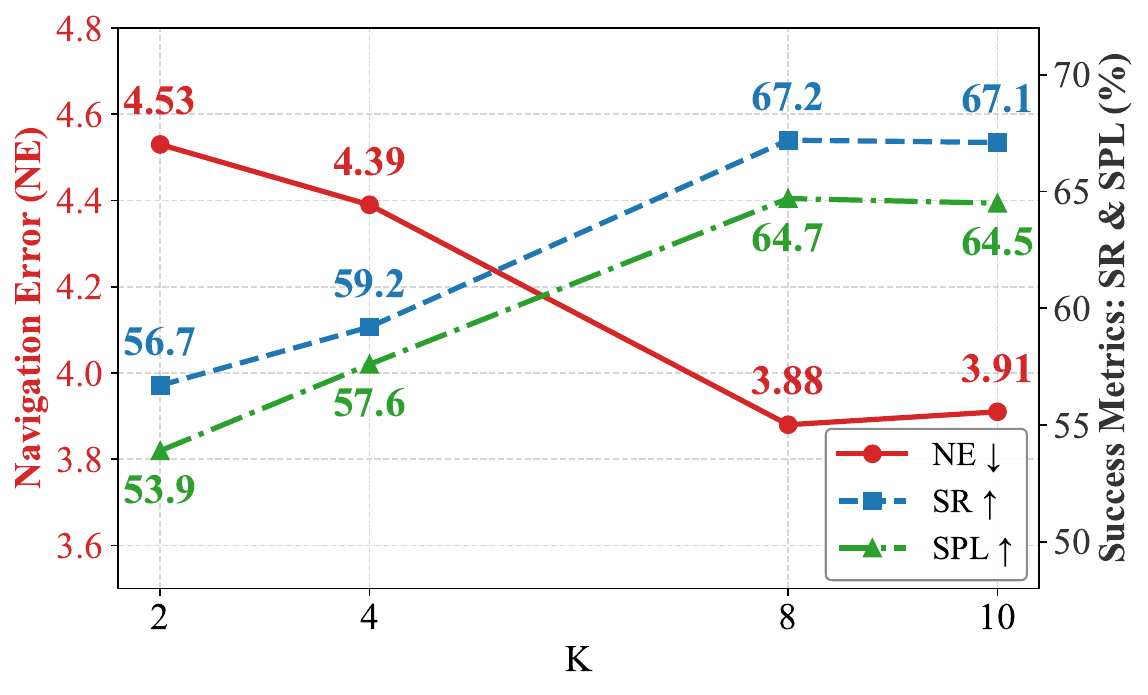}
  \caption{Ablation on the impact of different temporal context length.}
  \label{fig:context}
  \vspace{-16pt} 
\end{wrapfigure}

To rigorously investigate the comprehensive impact of context length on overall navigation performance, we conducted an ablation study evaluating temporal context windows of varying lengths on the R2R validation set. As shown in ~\cref{fig:context}, when the model is deprived of an effective temporal receptive field, the system inevitably succumbs to short-sightedness, yielding a mere 56.7\% SR. In complex indoor environments, an overly sparse history of frames fundamentally fails to provide the model with sufficient prior contextual information. 
The VLM struggles to build a coherent awareness of its motion trajectory, making it highly susceptible to disorientation at local dead-ends or instruction inflection points.
Conversely, when the context window is optimally expanded to 8 frames, the model successfully achieves precise 
alignment between high-level navigational reasoning and dynamic visual transformations, peaking at an SR of 67.2\% and SPL of 64.7\%. 
However, further extending the sequence length actively exacerbates the issue of attention dilution. An excessively long context window inadvertently introduces a substantial volume of stale visual features that are entirely irrelevant to the current decision-making step. This accumulated visual noise inevitably disperses the VLM's attention weights away from critical, real-time visual prompts, thereby inducing the model to generate severe spatial hallucinations and degrading overall trajectory planning.

\section{Conclusion}
 
In this paper, we present AgentVLN, a novel and efficient general embodied framework. Unlike conventional methods  that rely on massive datasets for implicit 3D representation learning, we introduce a VLM-as-Brain paradigm that decoupled high-level natural language cognitive reasoning from low-level perceptual control. This paradigm empowers the agent to dynamically adapt to unseen environments via a plug-and-play skill library. Specifically, we design a cross-space representation mapping mechanism that successfully eliminates the representation mismatches inherent in multi-level architectures. Synergized with a context-driven self-correction and active exploration strategy, this mechanism significantly bolsters the model's generalization performance. Furthermore, to tackle the persistent issue of inaccurate local target localization, we present the QD-PCoT, which effectively resolves semantic and spatial ambiguities within complex, unstructured environments while incurring zero parameter overhead. Finally, we construct and open-source AgentVLN-Instruct, a large-scale instruction-tuning dataset engineered with a dynamic stage-routing mechanism. Ultimately, our work establishes a robust and practical paradigm for the development of next-generation lightweight embodied navigation frameworks.

% ========================================================

% \section*{Acknowledgements}
% Please insert your acknowledgments here.

\bibliographystyle{splncs04}
\bibliography{main}
\end{document}